\documentclass[journal,twoside,web]{ieeecolor}
\usepackage{generic}
\usepackage{cite}
\usepackage{amsmath,amssymb,amsfonts}
\usepackage{graphicx}
\usepackage{algpseudocode}
\usepackage{algorithm}
\usepackage{hyperref}
\usepackage{booktabs}
\hypersetup{}
\usepackage{textcomp}
\def\BibTeX{{\rm B\kern-.05em{\sc i\kern-.025em b}\kern-.08em
    T\kern-.1667em\lower.7ex\hbox{E}\kern-.125emX}}

\providecommand{\refname}{References}


\begin{document}
\title{Ultrasound Operator Guidance Using World Modeling and Retrieval Based Action Planning}
\author{Noortje I.P. Schueler, Hans van Gorp, and Ruud J.G. van Sloun
\thanks{This work was supported by the Dutch Research Council (NWO) through the Research Program VIDI under Project 20381 (Corresponding author: Ruud J. G. van Sloun).}
\thanks{Noortje I.P. Schueler, Hans van Gorp, and Ruud J.G. van Sloun are with the Department of Electrical Engineering, Eindhoven University of Technology, 5612 AZ Eindhoven, The Netherlands (e-mail: n.i.p.schueler@tue.nl).}
}

\maketitle

\begin{abstract}
Ultrasound is widely used, but acquisition quality is heavily dependent on the operator's knowledge and expertise. With demand for examinations outpacing the supply of trained sonographers, operator-guidance systems aim to close this gap by instructing a less trained user how to move the probe toward a target view. 
In this paper, we propose a retrieval-induced latent transition model for ultrasound acquisition dynamics, formulating ultrasound operator guidance as multi-step planning and retrieval in a world model. Using a V-JEPA 2.1 backbone, observations are first encoded into a latent space where anatomically related views lie close together. We then retrieve similar views from a reference database containing encoded latent states and corresponding probe positions and orientations. Rather than learning a parametric transition function, we directly use physically executed transitions from the database to establish our nonparametric, retrieval-induced transition model that supports receding-horizon planning. At deployment, guidance is generated from the live ultrasound image feed alone, without any probe tracking hardware. Applied to carotid ultrasound, the proposed planner reaches the target view in 86\% of retrospective closed-loop episodes, versus 52\% and 43\% for representative baselines, outperforming both on every target view, including the challenging longitudinal internal and external carotid artery views. A prospective feasibility study on unseen volunteers, run in real time on a CPU using distillation, reaches 83\% target-view reachability. Because planning is driven by proximity to any encodable goal latent, the same world model can navigate back to any previously acquired, patient-specific frame, supporting reproducible longitudinal imaging for e.g. perioperative or follow-up monitoring.
\end{abstract}
\section{Introduction}
\label{sec:introduction}
\IEEEPARstart{U}{ltrasound} is one of the most widely used medical imaging modalities due to its real-time imaging capabilities, portability, safety, and low cost\cite{jiang2020automatic}. However, one downside is that ultrasound is inherently operator dependent, as image quality is determined not only by patient anatomy but also by probe positioning\cite{kojcev2017reproducibility, scorza2015comparative}. Acquiring clinically relevant views therefore requires substantial practical expertise, creating a barrier to consistent image acquisition and wider accessibility of ultrasound examinations.

At the same time, the demand for ultrasound examinations continues to grow while the number of trained sonographers remains limited\cite{won2024}. This imbalance has stimulated interest in workflows that enable image acquisition by less experienced operators while maintaining diagnostic quality. Achieving this goal requires reliable acquisition of standardized imaging views, as image interpretation and quantitative analysis depend critically on the quality and consistency of the acquired data. 

To address the challenges of ultrasound acquisition, autonomous robotic and human operator-guidance approaches have both been proposed. Robotic systems aim to remove the operator entirely, automating probe manipulation to standardize scanning. Operator-guidance systems instead instruct a human operator on how to manipulate the probe to reach a standardized target view. Despite this difference in execution, both approaches share the same underlying prediction problem: given the current observation, predict the probe motion that brings the current view closer to the target view. As we review in Section~\ref{sec:related}, existing methods differ chiefly in whether that predicted motion is evaluated against its consequences before being issued as guidance, and if so, how close to inference time that evaluation happens. Across the full range of prior work, however, that evaluation never extends beyond a single step.

Ultrasound acquisition, however, is a sequential navigation task under partial observation: ambiguous views such as vessel cross-sections do not uniquely determine location along the vessel, so an informative action may need to look several steps ahead rather than one. World models represent how observations evolve under actions, supporting exactly this kind of trajectory-level planning. Self-supervised training on large ultrasound datasets has already shown that a latent space can be learned that organizes anatomically related views\cite{echojepa}. Our key observation is that such a representation, combined with the scan trajectories used to train it, already implies a world model: rather than fitting a dynamics function, we retrieve observed latent transitions directly for planning. This extends the role of retrieval in ultrasound guidance from supplying context to a policy\cite{ragruss} to serving as the predictive model itself.

Building on this observation, we hypothesize that ultrasound operator guidance is better formulated as planning in a learned world representation than as direct action prediction. In our proposed method, observations are encoded with a video-JEPA (V-JEPA 2.1) backbone \cite{vjepa2_1}, candidate futures are obtained by vector search over previously observed transitions rather than by action regression, and retrieved trajectories are scored by their predicted likelihood of reaching the desired standard view and executed in a receding-horizon fashion, i.e., only the first action of the selected trajectory is issued, and the plan is recomputed from each new observation. Guidance therefore requires no pose sensing at inference; probe poses are needed only to build the trajectory database during training.

\begin{figure}[t]
    \centering
    \includegraphics[width=0.99\columnwidth]{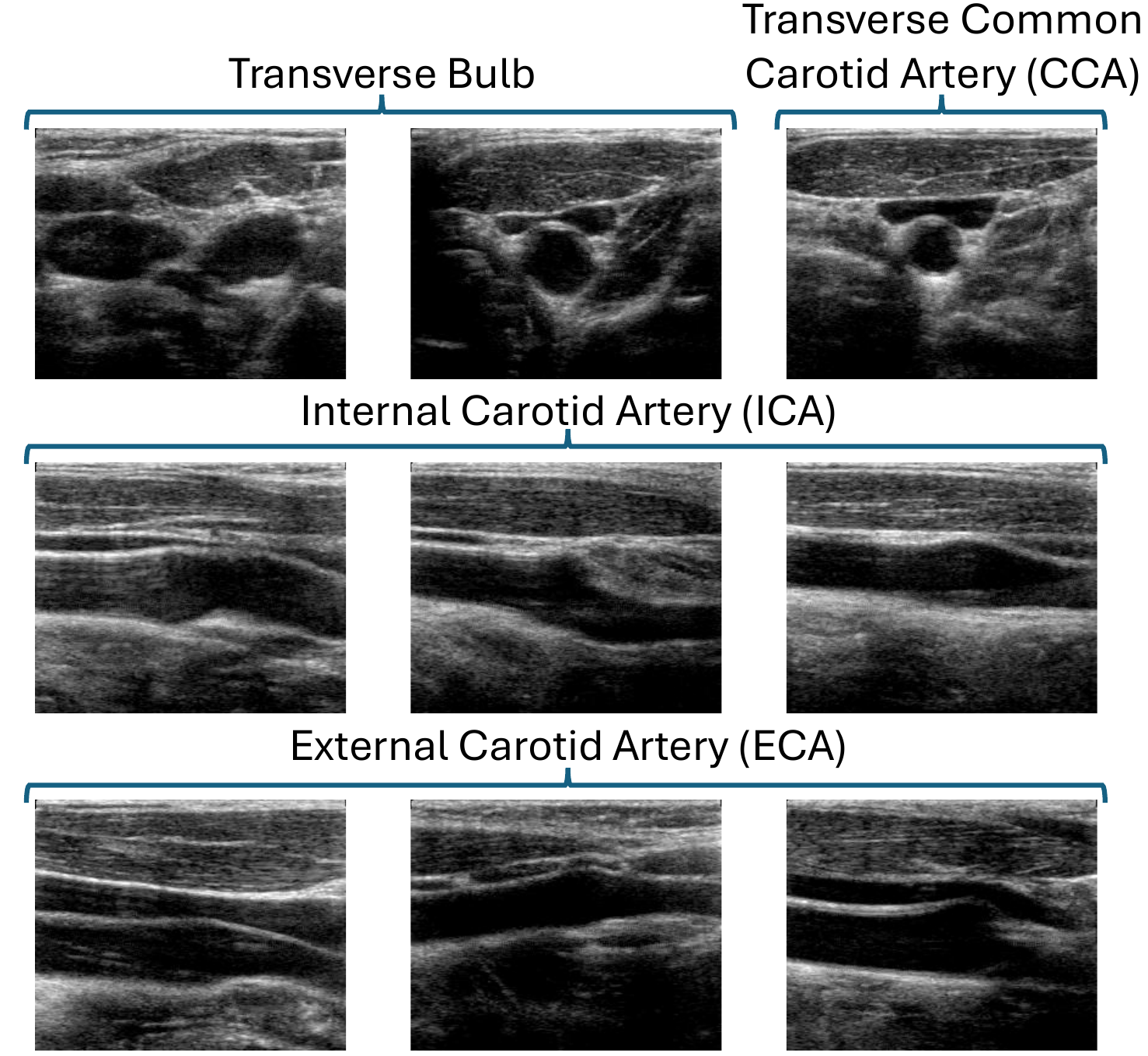}
    \caption{Navigation in carotid ultrasound is difficult for both human operators and AI guidance models alike, due to high intra-class variation and inter-class similarity between different ultrasound views. This figure illustrates both effects using representative images from different volunteers: intra-class variation, where the same anatomical view (transverse bulb, ICA, or ECA) appears visually different across volunteers, and inter-class similarity, where different anatomical views (e.g., transverse bulb (right image) and transverse CCA) can appear visually similar. \vspace{1em}}
    \label{fig:dataset}
\end{figure}

We evaluate this hypothesis in carotid ultrasound, a clinically important and technically demanding acquisition task. Accurate carotid assessment requires reproducible localization of vascular structures and Doppler sampling locations, while image appearance varies substantially across subjects and pathologies, as shown in Fig. \ref{fig:dataset}. Carotid navigation also demands full six-degree-of-freedom probe motion, unlike the rotation-only settings addressed by much of the prior literature. These properties make it a stringent testbed for world-model-based planning and robust operator guidance in ultrasound acquisition.

The main contributions of this work are as follows:
\begin{itemize}
\item \textbf{Guidance as multi-step planning.} We treat ultrasound guidance as planning several steps ahead, not just predicting a single action. At each step, we compare candidate trajectories in a learned latent space, pick the best one, and replan when a new observation comes in. 
\item \textbf{Retrieval as a nonparametric transition model.} Instead of training a parametric transition model, we use transitions recorded from real scans to build a nonparametric retrieval-induced transition model.
\item \textbf{Systematic evaluation against representative baselines.} We compare our method against one-shot action prediction and single-step planning baselines under the same evaluation setup. 

\item \textbf{Real-time prospective evaluation.} Beyond retrospective simulation, we run the pipeline in real time on a CPU in a prospective study on previously unseen volunteers, reaching 83\% target-view reachability.
\end{itemize}

\begin{figure*}[hbt!]
    \centering
    \includegraphics[width=0.99\linewidth, trim=0cm 2.2cm 0cm 0cm, clip]{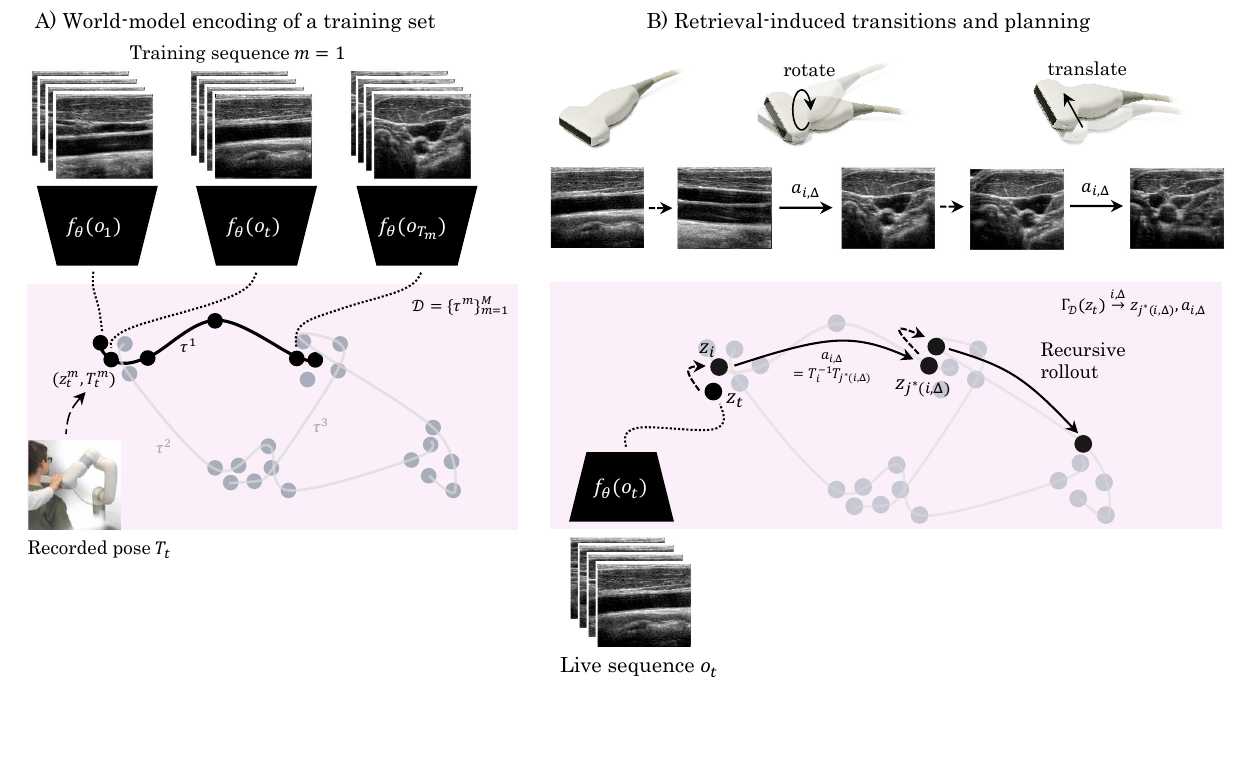}
    \caption{Visual overview of the ultrasound world model and planning in its latent space. A) Training sequences are encoded into latent states using the learned encoder $f_\theta$, based purely on the visual ultrasound frames; each training frame additionally has a ground-truth recorded pose, used to calculate actions between training frames. B) The live imaging sequence is encoded with the same encoder $f_\theta$. Many candidate trajectories are then randomly sampled by alternating between retrieval hops and within-trajectory hops. Retrieval hops sample the top-$K$ nearest neighbors of the current latent state to create the anchor points $z^{(h)}$. Within-trajectory hops then follow a training sequence forward or backward in time from anchor $z^{(h)}$ to $\bar z^{(h)}$, yielding a recorded action $a^{(h)}$. Here, we show one example of a sampled trajectory. From all generated trajectories, the lowest cost trajectory is chosen, and its first action is communicated to the user.
    }
    \label{fig:methods}
    \vspace{-1em}
\end{figure*}
\section{Related Work}
\label{sec:related}
Prior guidance and autonomous-scanning systems can be organized along the axis introduced above: whether a predicted action's consequences are evaluated before it is issued as guidance, and if so, how close to inference time that happens.

\textit{One-shot action prediction.} In the largest group, the prediction is the guidance. US-GuideNet couples a convolutional frame encoder with a recurrent model over orientations measured by an Inertial Measurement Unit (IMU) and predicts either the rotation needed to reach a standard plane or the rotation an expert would perform next; in both cases the rotation is emitted directly, with no translation guidance\cite{US-GuideNet}. The approach of Bao \textit{et al.} likewise pairs a frame encoder with a per-target head that predicts rotation only\cite{bao2024real}. UltraSeP adds masked pretraining over joint image-pose sequences, but the deployed model remains a one-shot regressor that uses probe pose\cite{ultrasep}. Autonomous carotid systems extend the same principle from a single target view to a full examination: UltraBot clones expert trajectories through policies trained for individual target views, inheriting expert behavior rather than improving on it\cite{ultrabot}, while UltraHiT decomposes carotid anatomy into a standard structure and individual variation, using a hierarchical gate to decide whether the current scan follows a known anatomical pathway or requires one of twelve learned corrective motions\cite{ultrahit}. Reliance on robot proprioception is reasonable in these autonomous settings, where the probe pose is readily available from position sensors. However, probe pose information is usually unavailable during freehand scanning, limiting the transfer of these methods to operator-guidance scenarios. RAG-RUSS closes this gap by returning to the operator-guidance setting at the scale of a full examination: rather than relying on probe pose, it retrieves relevant prior context to condition a language model that performs a full carotid examination in clinical-workflow order, explaining its current stage and next planned action\cite{ragruss}. Across all of these methods, nothing intervenes between the predicted quantity and the instruction issued: the model is never asked what the world would look like afterward.

\textit{Amortized consequence modeling.} A second group weighs the consequences of actions during training, but during deployment that reasoning has been amortized away. For example, VesNet-RL internalizes a long-horizon reward into a value function, though it navigates from binary vessel masks, discarding the appearance cues expert operators rely on, and restricts actions to in-plane translation and axial rotation\cite{vesnet}. EchoWorld is the operator-guidance counterpart of the same pattern: it uses predictive objectives during pretraining to learn anatomy- and motion-aware representations. During inference, however, guidance is still generated by direct action regression from encoded observations and motion history. World modeling shapes the representation here, but it is not explicitly used for planning\cite{echoworld}. Fan \textit{et al.} take the same route with a frozen latent-diffusion ultrasound model whose imagined rollouts supply a training-time reward for a goal-conditioned probe policy, so the world model shapes the policy but is never queried when guidance is issued\cite{fan2026action}.

\textit{Single-step inference-time planning.} Cardiac Copilot is the closest prior work to ours, and the only system we are aware of in which a forward model is genuinely queried during inference\cite{cardiaccopilot}. To do so, it uses an action-conditioned latent transition model called Cardiac Dreamer. The policy proposes a movement, the Dreamer imagines the latent state that movement would produce, and the action is refined in light of it. Two limitations remain, though. The approach is explicitly target-oriented, predicting the relation to the target ``without the need to consider the intermediate path''\cite{cardiaccopilot}, so the lookahead is a single refinement step rather than optimization over a trajectory. Moreover, goals are restricted to a fixed set of predefined target embeddings.

\textit{Retrieval-based and non-parametric planning beyond ultrasound.}
Non-parametric reuse of previously collected experience has a long history in visual control and planning, with methods differing in what is retrieved and how retrieved experience is converted into actions. At the shortest horizon, VINN predicts each action through locally weighted regression over the (k) nearest demonstrations in a learned visual embedding space, without explicitly reasoning over future consequences \cite{pari2021surprising}. More recent retrieval-as-control approaches such as RT-Cache retrieve a short action sequence associated with the most similar stored observation, execute this sequence for several steps, and then re-query its memory from the newly observed state \cite{kwon2025rt}. Longer-horizon methods instead use stored experience to construct intermediate goals. SPTM, SoRB, and ViNG organize previously observed visual states into topological memories or graphs and search these structures for subgoals, while a separate local control mechanism converts each selected visual subgoal into executable actions \cite{savinov2018semi, eysenbach2019search, shah2021ving}. Most closely related, PALMER retrieves previously experienced trajectory segments between nearby states and repurposes sampling-based planning and graph search to combine these segments into approximately optimal start-to-goal paths. Execution of the planned state trajectory again relies on a learned goal-conditioned controller \cite{beker2022palmer}. Other approaches use retrieval and trajectory recombination primarily offline: model-based trajectory stitching connects states from different demonstrations using model-generated actions to construct improved behavioral-cloning trajectories \cite{hepburn2024model}, while Graph Search and Retrieval uses graph search over offline experience to identify desirable behaviors that are subsequently distilled into a parametric imitation policy \cite{yin2024offline}.

\textit{Multi-step world-model planning beyond ultrasound.}
A complementary line of work performs explicit multi-step planning by learning a parametric model of action consequences. Visual-foresight approaches recursively predict the outcomes of candidate actions and optimize action sequences within a model-predictive-control framework \cite{finn2017deep,ebert2018visual}. More recently, DINO-WM learns action-conditioned dynamics in a pretrained DINOv2 representation space and optimizes action sequences at inference by comparing recursively predicted future latent states with a visual goal \cite{zhou2024dino}. V-JEPA 2-AC similarly learns an action-conditioned latent predictor on top of a pretrained visual representation and uses predicted latent rollouts for image-goal model-predictive control \cite{vjepa2_1}. These approaches demonstrate that long-horizon visual planning can be performed directly in learned representation spaces, but rely on a learned transition function to generate the hypothetical states considered during planning.\newline
We combine both perspectives by planning over the consequences of candidate actions, as world-model planners do, while retrieving these consequences from observed experience rather than generating them with a learned dynamics model. \newline

Across prior ultrasound guidance methods, action consequences are incorporated increasingly close to inference time: they are absent from purely reactive guidance, modeled during training in world-model-assisted policy approaches, and used for a one-step action refinement in Cardiac Copilot. However, these methods do not explicitly compare multiple action sequences and their longer-horizon consequences at deployment. Our approach combines the inference-time planning principle of general-purpose world models with the non-parametric reuse of recorded experience. Rather than immediately replaying retrieved behavior, searching stored experience for a complete route to the goal, or recursively generating future states with a learned transition model, we use retrieval itself as the forward transition operator of an online planner. From the current ultrasound state, nearby recorded states and their observed outgoing state-motion-state transitions are retrieved and recursively expanded into a tree of finite-horizon futures. These branches are evaluated relative to the desired target view and probe-motion regularity. Each transition supplies both an empirically observed successor state and the recorded six-degree-of-freedom probe displacement that produced it; only the first displacement of the selected trajectory is communicated to the operator before replanning from the newly acquired image. Retrieval therefore serves not merely as a source of actions, trajectory segments, or intermediate subgoals, but as a non-parametric action-conditioned transition model for receding-horizon ultrasound planning.

\section{Method}
\label{sec:method}
\subsection{General setup}
We formulate ultrasound operator guidance as a navigation problem in a learned imaging world. Given a live ultrasound observation $o_t$ and a desired target view $v_g\in\mathcal{V}$, the label set of standard clinical views defined in Section~\ref{sec:dataset_details}, the goal is to generate guidance instructions that move the operator from the current anatomical location toward the target imaging plane. Unlike existing guidance systems that directly predict the next probe action, we perform multi-step planning in a latent anatomical state space. See Fig.~\ref{fig:methods} for a visual overview of the method.

Observations are first mapped to latent states, where nearby states correspond to anatomically related ultrasound views. Previously acquired scan trajectories are used to construct a database of states and probe locations. Rather than learning an explicit dynamics model, guidance is generated by retrieving and scoring candidate trajectories from this database and communicating the first action of the best one to the operator, repeated after every new observation in a receding-horizon fashion (Section~\ref{sec:guidance_generation}).

\subsection{Dataset and implementation details}
\label{sec:dataset_details}
Ultrasound data were acquired using a tracked freehand carotid ultrasound protocol: the probe was mounted on a robotic arm that a human operator manipulated freehand, while the arm's encoders provided precise, synchronized six-degree-of-freedom probe poses alongside each recorded B-mode frame. This tracking setup is required only to build the training-time transition database; the guidance system itself needs no pose sensing once deployed (Section~\ref{sec:introduction}). Each acquisition therefore provides both a sequence of ultrasound observations and a sequence of probe poses, from which image transitions and navigation trajectories can be extracted.

The six degrees of freedom of the probe pose consist of its translation, in millimeters, relative to a reference origin, and its orientation, expressed as pitch, roll, and yaw in degrees, relative to a base orientation. Both the origin and the base orientation are arbitrary, since the method uses only relative displacements between poses, never their absolute values. We transform each recorded pose into a homogeneous transformation matrix
\begin{equation}
\label{eq:transformation_matrix}
    T_t=\begin{bmatrix} R & \vec{p} \\ 0 & 1 \end{bmatrix} \in SE(3),
\end{equation}
where $\vec{p}\in\mathbb{R}^3$ is the translation vector, $R\in SO(3)$ is the rotation matrix corresponding to the recorded pitch, roll, and yaw, and the subscript $t$ indexes acquisition time, consistent with its use elsewhere in the paper. Relative probe motions used throughout the framework are then obtained by composing and inverting these matrices, rather than by operating on the Euler angles directly.

The dataset consists of 48 recorded trajectories, each from a different volunteer who gave informed consent, totaling 247{,}827 ultrasound frames. Data were partitioned at the subject level into training ($n=31$), validation ($n=4$), and test ($n=13$) sets to prevent information leakage between splits.

The dataset was manually annotated with clinically relevant carotid views: short-axis common carotid artery (CCA), bifurcation, carotid bulb, longitudinal CCA, internal carotid artery (ICA), and external carotid artery (ECA). In-between frames that did not belong to one of the standard views were labeled ``no view'', allowing the view-classification head to explicitly reject frames that do not correspond to any standard view. The six standard views plus the ``no view'' class constitute the label set $\mathcal{V}$ used throughout the paper. All frames in the validation and test splits were labeled in this way, whereas only a fraction of the training split, \(2.6\%\), was labeled.

\subsection{Latent ultrasound world modeling}
\label{sec:latent_representation}
Navigation performance depends critically on the quality of the learned latent representation. To support planning, the latent space should preserve both anatomical similarity and probe-motion dynamics, enabling retrieval of visually similar observations while maintaining meaningful state transitions.

We build upon V-JEPA 2.1 and retrain its backbone on carotid ultrasound video sequences. The model receives a clip of four consecutive ultrasound frames and produces two latent tubelet representations, each spanning two consecutive frames, a consequence of V-JEPA 2.1's own spatiotemporal patchification rather than a choice made in this work. Because guidance decisions should be based on the current view, only the final tubelet, corresponding to the most recent two frames, is retained. Each observation $o_t$ therefore denotes a four-frame clip ending at time $t$, rather than a single image.

The selected tubelet tokens are then aggregated into a compact global state representation by an attentive pooling module. We denote the fine-tuned V-JEPA 2.1 backbone described above by $h_\theta$, and this pooling module by $g_\theta^{\text{pool}}$; both are learned, so $\theta$ collectively denotes the parameters of the backbone and this pooling module together. The full encoder is their composition,
\begin{equation}
\label{eq:attentive_pooling}
z_t = f_\theta(o_t) := g_\theta^{\text{pool}}\bigl(h_\theta(o_t)\bigr) \in \mathbb{R}^d,
\end{equation}
and it is this composite $f_\theta$ that we invoke throughout the remainder of the paper whenever an observation is mapped to its latent state for retrieval and planning.

To shape this representation, three lightweight heads are attached on top of the pooled state during training only, and discarded afterward: an auxiliary view-classification head, predicting a distribution over the label set $\mathcal{V}$ via softmax; and an inverse-dynamics head and a forward-dynamics head, additionally predicting, respectively, the relative action between two latents and the latent reached by applying a given action,
\begin{equation}
\label{eq:dynamics_heads}
\hat a = g_\theta^{\text{action}}(z,z'), \qquad \hat z' = g_\theta^{\text{forward}}(z,a).
\end{equation}
All three heads serve only to shape the representation during training; none of their predictions are used afterward. In particular, planning remains fully nonparametric and relies exclusively on the recorded actions stored in the transition database (Eq.~\ref{eq:relative-action}), never on $g_\theta^{\text{action}}$ or $g_\theta^{\text{forward}}$ predictions, and the view probabilities used for goal specification come instead from a separate, dedicated module, described next.

Because accurate view classification matters for guidance, we train a second, independently parameterized attentive pooling module on top of the same frozen backbone $h_\theta$, together with its own view-classification head; we denote their parameters collectively by $\psi$. This yields
\begin{equation}
\label{eq:view_head}
p(o) := g_\psi^{\text{view}}\bigl(g_\psi^{\text{pool}}(h_\theta(o))\bigr),
\end{equation}
with $p_v(o)$ denoting the probability assigned to view $v\in\mathcal{V}$. Trained solely for this task, rather than jointly with the shaping heads above, this dedicated branch achieves higher view-classification accuracy, and it is $p$ that is used throughout the remainder of the paper for goal specification and trajectory scoring, never the auxiliary head of the previous paragraph. Every observation stored in or queried against the transition database has both its latent state $z_t$ (Eq.~\ref{eq:attentive_pooling}) and its view probability $p(o_t)$ computed and cached together, so we write $p(z_t)$ as shorthand for the view probability associated with state $z_t$.

Training proceeds in three stages, all on the training split of our ultrasound dataset. We first retrain $h_\theta$ from V-JEPA 2.1's released pretrained weights (from the base model size), following V-JEPA 2.1's own training paradigm; $h_\theta$ is then frozen. On top of this frozen backbone, we then independently train two branches, following the attentive-pooling recipe released with V-JEPA 2.1: the pooling module $g_\theta^{\text{pool}}$ jointly with the three shaping heads above, and, separately, the dedicated pooling module and view-classification head parameterized by $\psi$. Both branches are trained on the same training split; the view-classification heads are supervised on the subset of frames with view labels, while the two dynamics heads require no labels and are therefore trained on the entire split. The latent state $z_t$ from the first branch serves as the basic unit for retrieval, planning, and navigation, while the view probability $p$ from the second branch guides planning toward the desired view, as detailed in Section~\ref{sec:trajectory_scoring}.

Every state in the transition database is assigned a view-membership probability $p(z)$ (Eq.~\ref{eq:view_head}). Where the underlying frame has a manual view label, we use the corresponding ground-truth one-hot vector directly rather than the classifier's prediction; the dedicated view-classification head is applied only to states whose frame lacks a manual label. This distinction was only necessary for the training split, since the validation and test splits were fully labeled (Section~\ref{sec:dataset_details}). The result is a per-state view probability, soft or ground-truth one-hot, throughout the transition database, which is what lets planning be guided toward a desired standard view without requiring exemplar goal images for that view, as detailed in the following.

\subsection{Receding-horizon planning}
\subsubsection{Retrieval-induced latent transition operator}
Our goal is to establish a nonparametric retrieval-induced latent transition operator that maps latent states to realistic action-conditional future states. Let the historical database contain \(M\) acquisitions,
\begin{equation}
\mathcal D = \left\{ \tau^m \right\}_{m=1}^{M},
\end{equation}
where
\begin{equation}
\tau^m = \left( z_1^m,T_1^m, z_2^m,T_2^m, \ldots, z_{T_m}^m,T_{T_m}^m \right),
\end{equation}
and \(z_s^m\in\mathbb{R}^d\) is the latent representation of frame \(s\) in acquisition \(m\), and \(T_s^m\in SE(3)\) is the corresponding tracked probe pose. We write $\tau^m$ for a recorded acquisition and, with a slight abuse of notation, $\tau$ (without a superscript) for a candidate trajectory assembled at planning time (Section~\ref{sec:trajectory_scoring}); unlike a $\tau^m$, a candidate trajectory need not stay within a single acquisition. For notational convenience, we define the set of database-state indices as
\begin{equation}
\mathcal I = \left\{ (m,s) \;\middle|\; 1\leq m\leq M,\; 1\leq s\leq T_m \right\},
\end{equation}
such that for each index \(i=(m,s)\in\mathcal I\), we have shorthand notation
\begin{equation}
z_i=z_s^m, \qquad T_i=T_s^m.
\end{equation}
To define the latent transition operator acting on latent state \(z\), we first retrieve its \(K\)=10 nearest database states according to cosine similarity:
\begin{equation}
\mathcal N_K(z) = \operatorname{TopK}_{i\in\mathcal I} \operatorname{cos}(z,z_i).
\label{eq:latent-retrieval}
\end{equation}
These retrieved database states act as anchors from which physically observed transitions are sampled. For each retrieved anchor \(i=(m,s)\in\mathcal N_K(z)\) and frame stride \(\Delta\in\mathcal S(z)\)\footnote{Note that for computational tractability, we select a limited frame stride set \(\Delta\in\mathcal S(z)\), which contains $N_S$ steps with a stepsize that is adaptively determined based on the target-view probability $p_{v_g}(z)$: higher probability yields smaller strides (refinement), whereas low probability bigger steps.}, the planner considers the forward and backward successor indices
\begin{equation}
i^{+} = (m,s+\Delta), \qquad i^{-} = (m,s-\Delta),
\label{eq:forward-backward-candidates}
\end{equation}
and first selects the forward or backward successor $j\in\{i^{-},i^{+}\}$ with the highest immediate target-view probability,
\begin{equation}
j^\star(i,\Delta) = \operatorname*{arg\,max}_{j\in\{i^{-},i^{+}\}} p_{v_g}(z_j),
\label{eq:successor-selection}
\end{equation}
having associated relative probe motion:
\begin{equation}
a_{i,\Delta} = T_i^{-1}T_{j^\star(i,\Delta)}.
\label{eq:relative-action}
\end{equation}
The adaptive retrieval-induced latent transition operator is then the set-valued one-to-many map containing actions and future states: 
\begin{equation}
\Gamma_{\mathcal D} : \mathbb{R}^d \rightrightarrows \mathcal A \times \mathbb{R}^d,
\end{equation}
defined as
\begin{equation}
\Gamma_{\mathcal D}(z) = \left\{ \left( a_{i,\Delta}, z_{j^\star(i,\Delta)} \right) \;\middle|\; i\in\mathcal N_K(z),\; \Delta\in\mathcal S(z) \right\}.
\label{eq:adaptive-transition-operator}
\end{equation}
Each candidate transition can be interpreted as the nonparametric model:
\begin{equation}
z \;\dashrightarrow\; z_i \;\xrightarrow{\,a_{i,\Delta}\,}\; z_{j^\star(i,\Delta)},
\end{equation}
where the dashed arrow denotes retrieval of a similar database state and the solid arrow denotes a corresponding recorded probe motion in the database.

\subsubsection{Recursive trajectory expansion}
\label{sec:guidance_generation}
At inference time, the current ultrasound observation \(o_t\) is encoded as
\begin{equation}
z_t = f_\theta(o_t)
\end{equation}
(Eq.~\ref{eq:attentive_pooling}). Starting from $\bar z^{(0)} := z_t$, the latent transition operator is applied recursively to generate candidate trajectories: at each planning step $h=1,\ldots,H$,
\begin{equation}
\bigl( a^{(h)}, \bar z^{(h)}, c^{(h)} \bigr) \in \Gamma_{\mathcal D}\bigl(\bar z^{(h-1)}\bigr),
\label{eq:recursive-planning}
\end{equation}
where we in practice choose a horizon \(H=6\), and where the retrieved anchor underlying each transition is denoted $z^{(h)}$.

After each transition, the hop target \(\bar z^{(h)}\) becomes the query state for the next expansion. The planner therefore retrieves a new set of nearest neighbors at every planning depth. This gives the recursive structure
\begin{equation}
\begin{aligned}
\bar z^{(h-1)} &\;\dashrightarrow\; z^{(h)} \;\xrightarrow{\,a^{(h)}\,}\; \bar z^{(h)} \\
&\;\dashrightarrow\; z^{(h+1)} \;\xrightarrow{\,a^{(h+1)}\,}\; \bar z^{(h+1)}.
\end{aligned}
\end{equation}
Because a new retrieval is performed after every transition, consecutive actions may originate from different acquisitions. This allows the planner to combine locally compatible recorded motions into trajectories that are not explicitly present in the database (see also the example in Fig. \ref{fig:methods}).
A candidate trajectory of horizon \(H\) is written as
\begin{equation}
\tau = \bigl\{\bigl( \bar z^{(h)},a^{(h)} \bigr)\bigr\}_{h=1}^{H},
\end{equation}
with action sequence
\begin{equation}
a(\tau) = \left( a^{(1)},a^{(2)},\ldots,a^{(H)} \right).
\end{equation}
Recursive application of \(\Gamma_{\mathcal D}\) produces a branching search tree whose size would otherwise grow exponentially with the horizon \(H\). To keep this tractable, candidate trajectories are grouped into lanes according to the initial retrieved anchor \(i\in\mathcal N_K(z_t)\) from which they originated at the first planning step, and each lane independently retains at most \(N_{\mathrm{lane}}=5\) trajectories. Every \(H_{\mathrm{collapse}}=2\) expansion steps, that is, after every even planning step, the trajectories within each lane are pruned back down to the best \(N_{\mathrm{lane}}\) according to the per-state target-view cost of their current frontier state, \(d_h=1-p_{v_g}\bigl(\bar z^{(h)}\bigr)\) (Section~\ref{sec:trajectory_scoring}), rather than the full trajectory cost \(J(\tau)\). Because \(d_h\) is obtained directly from the cached, per-state view probability already stored for every state in the transition database (Section~\ref{sec:latent_representation}), whether a ground-truth one-hot label or a soft classifier prediction, this pruning reduces to a database lookup and is substantially cheaper than evaluating the full cost function, whose remaining terms require the entire trajectory history rather than a single state. This periodic, cheap pruning step keeps the search tree tractable while preserving several plausible continuations per lane.

Let \(\mathcal T(z_t)\) denote the set of trajectories retained after the final expansion. The optimal trajectory is selected as
\begin{equation}
\tau^\star = \operatorname*{arg\,min}_{\tau\in\mathcal T(z_t)} J(\tau),
\label{eq:optimal-trajectory}
\end{equation}
where \(J(\tau)\) is the trajectory cost function (Section~\ref{sec:trajectory_scoring}). Following a receding-horizon strategy, only the first action of the selected trajectory is communicated to the user,
\begin{equation}
a_t = a^{(1)}(\tau^\star).
\end{equation}
A new ultrasound observation is then acquired, and the complete planning procedure is repeated.

\subsubsection{Trajectory scoring}
\label{sec:trajectory_scoring}
Candidate trajectories are ranked using a cost function that balances target-view likelihood, consistent progress toward the target, and smooth probe motion:
\begin{equation}
\label{eq:trajectory_cost}
J(\tau)=
\lambda_{\mathrm{traj}} J_{\mathrm{traj}}
+
\lambda_{\mathrm{mom}} J_{\mathrm{mom}}
+
\lambda_{\mathrm{xstep}} J_{\mathrm{xstep}},
\end{equation}
where \(\lambda_{\mathrm{x}}\) and \(J_{\mathrm{x}}\) are cost weights and cost components, respectively.
For a candidate trajectory:
\[
\tau=\left\{\left(\bar z^{(h)},a^{(h)}\right)\right\}_{h=1}^{H},
\]
we define the per-state target-view cost as
\[
d_h = 1-p_{v_g}\!\left(\bar z^{(h)}\right),
\]
where
\(p_{v_g}(\bar z^{(h)})\)
is the probability assigned by the frozen view-classification head to the desired target view
\(v_g\in\mathcal{V}\).
A lower value of \(d_h\) therefore indicates greater confidence that the latent state corresponds to the target view.
Each candidate trajectory is scored by its average target-view cost over the planning horizon,
\[
J_{\mathrm{traj}}
=
\frac{1}{H}
\sum_{h=1}^{H}
d_h,
\]
which favors trajectories that consistently remain close to the target-view manifold throughout the rollout.
The remaining terms regularize the predicted probe motion.

The momentum term \(J_{\mathrm{mom}}\) penalizes changes in direction between consecutive actions within a candidate trajectory, encouraging smooth motion. The cross-step term \(J_{\mathrm{xstep}}\) penalizes reversals between the first action in the trajectory and the action executed at the previous planning step, reducing oscillatory behavior across replanning iterations. Both penalties are computed as \(1-\cos(\cdot,\cdot)\) between the dominant translation and rotation axes of the corresponding six-dimensional action vectors after rescaling translation and rotation to a common scale.
The planner selects the candidate trajectory with the lowest total cost \(J(\tau)\).

\section{Results}
\subsection{Latent space analysis}
To evaluate the quality of the learned state representation described in Section~\ref{sec:latent_representation}, we assess three complementary properties that are essential for planning: 1) anatomical separability, 2) local retrieval consistency, and 3) correspondence with probe-motion dynamics. We do so by comparing the proposed V-JEPA 2.1 backbone for the encoder with two possible alternative world models: EchoJEPA \cite{echojepa} and ImageJEPA \cite{imagejepa}. Lastly, to verify that the latent space gets organized by view, but generalizes over individual patients, we do a qualitative analysis of the embedding space using T-SNE \cite{tsne}.

\subsubsection{Linear probing}
We perform a simple linear probe on the proposed V-JEPA 2.1 backbone and the two baselines for view classification. Table \ref{tab:view_probe} reports the classification accuracy and the F1 scores. As we can see, the proposed encoder outperforms the other two baselines.

\subsubsection{Local retrieval consistency}
Since planning relies on nearest-neighbor retrieval, we additionally evaluate whether local neighborhoods in the latent space are anatomically consistent. For each labeled query frame, the ten nearest neighbors are retrieved from the test set using cosine similarity in the learned state space. We then calculate the retrieval purity as the fraction of retrieved frames that have the same label. These results are also reported in Table \ref{tab:view_probe}, where we can see that again the proposed method is more likely than the baselines to retrieve a latent code from a frame with an identical label as the query. 

\subsubsection{Transition consistency}
To evaluate whether the learned state representation also preserves probe-motion dynamics, we compute the cosine distance between pairs of latent states originating from the same acquisition and compare these distances with the magnitude of the corresponding recorded probe motions. The relationship between latent and physical distances is quantified using both the Pearson and Spearman correlation coefficients. Pearson correlation measures the degree of linear correspondence between latent and physical distances, whereas Spearman correlation evaluates whether their monotonic ordering is preserved. Together, these metrics assess whether larger probe motions correspond to larger displacements in the learned state space and whether this relationship is approximately linear. The results of this analysis are shown in Table \ref{tab:transition_dynamics}, all encoders show comparable correlations between latent distance and translational probe motion, whereas for rotation V-JEPA 2.1 shows the strongest correlation.

\begin{table}[t]
\centering
\caption{Latent Space Analysis.}
\begin{tabular}{lccc}
\toprule
& \multicolumn{2}{c}{Linear probe} & \multicolumn{1}{c}{Top-K} \\
\cmidrule(lr){2-3} \cmidrule(l){4-4}
Method &
Balanced accuracy &
Macro F1 &
Purity@10 \\
\midrule
EchoJepa & 0.75 & 0.75 & 0.65 \\
ImageJepa & 0.52 & 0.51 & 0.49 \\
V-JEPA 2.1 & \textbf{0.81} & \textbf{0.81} & \textbf{0.69} \\
\bottomrule
\end{tabular}
\label{tab:view_probe}
\end{table}
\begin{table}[t]
\centering
\caption{Representation dynamics results.}
\begin{tabular}{lcccc}
\toprule
& \multicolumn{2}{c}{Translation dynamics} &
  \multicolumn{2}{c}{Rotation dynamics} \\
\cmidrule(lr){2-3}
\cmidrule(lr){4-5}
Method &
Pearson &
Spearman &
Pearson &
Spearman \\
\midrule
EchoJEPA  & 0.20 & \textbf{0.16} & 0.30 & 0.34 \\
ImageJEPA & 0.19 & 0.15 & 0.31 & 0.33 \\
V-JEPA 2.1  & 0.20 & 0.13 & \textbf{0.41} & \textbf{0.47} \\
\bottomrule
\end{tabular}
\label{tab:transition_dynamics}
\vspace{-1.5em}
\end{table}

\subsubsection{Latent visualization}
\begin{figure}[t]
    \centering
    \includegraphics[width=0.99\columnwidth]{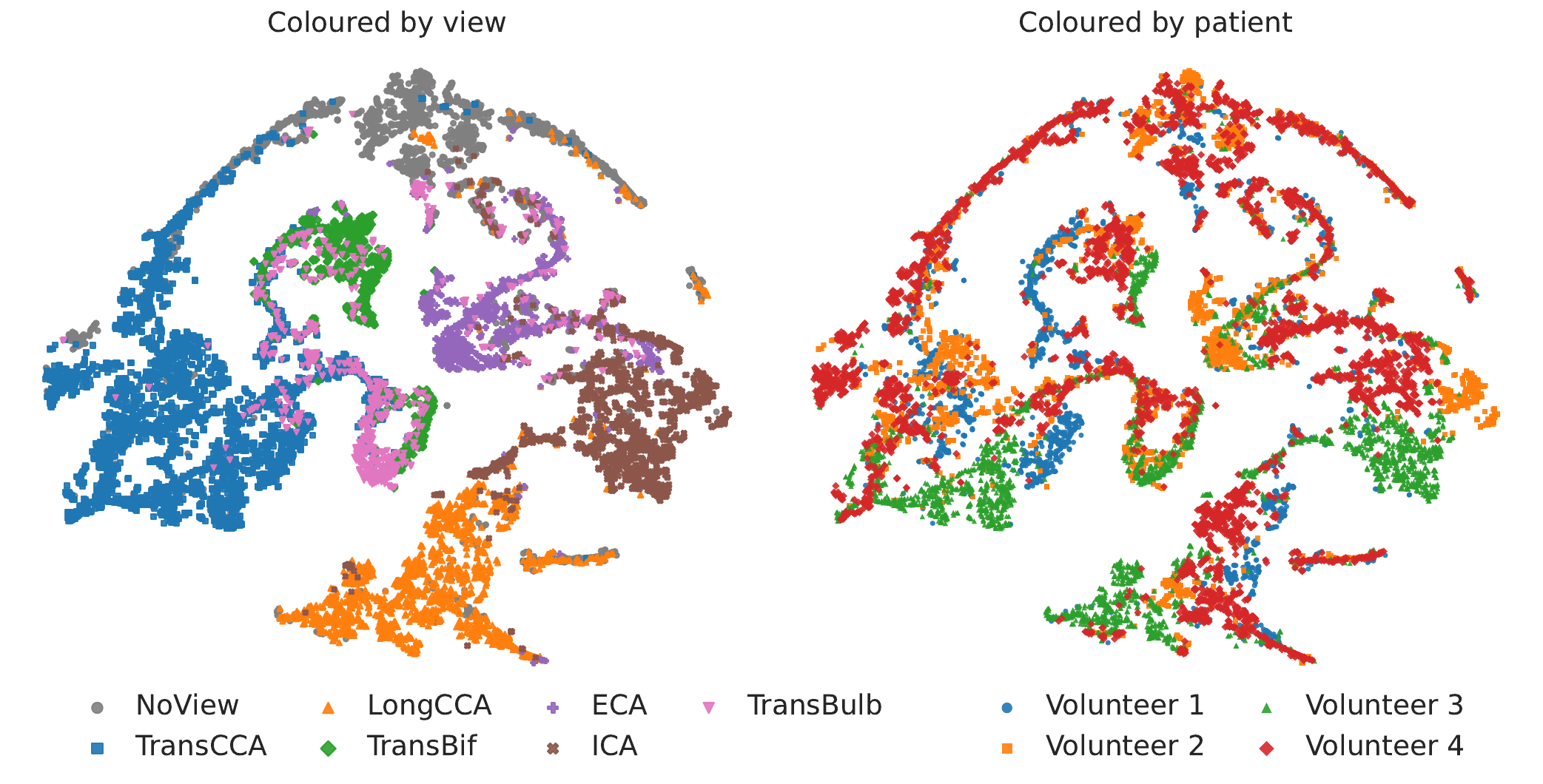}
    \caption{2D T-SNE embedding of the latent space of the validation set, colored by both view label and volunteer. Clear clusters appear over views, while the method generalizes neatly across volunteers.}
    \vspace{-1em}
    \label{fig:representation}
\end{figure}

We perform a T-SNE embedding on the latent codes of the validation set, using the cosine distance as the distance metric (since we use it also for Top-K retrieval). The resulting 2D embedding is shown in Fig. \ref{fig:representation}. Where we show the same embedding twice: colored by view and by volunteer. It can be seen from Fig. \ref{fig:representation} that the embedding are clearly clustered by view, with `bridges' appearing between similar views. At the same time, no such clustering appears with respect to the volunteers, clearly indicating that the latent encoder has generalized over volunteers.

\subsection{Retrospective navigation evaluation}
We evaluate navigation performance through a retrospective closed-loop simulation on held-out acquisitions. Each held-out acquisition is a densely sampled sequence of ultrasound images with corresponding tracked probe poses, so the recorded trajectory acts as a surrogate environment in which probe motions can be replayed against real observations. Every frame in the test set carries a view label drawn from the label set $\mathcal{V}$ (Section~\ref{sec:dataset_details}), spanning the six standard planes and the background \emph{no view} class.

At each planning iteration, the current live image is encoded into a latent state $z_k$ (Eq.~\ref{eq:attentive_pooling}), paired with the tracked probe pose $T_k$ of that frame. The planner predicts a relative probe motion $\hat{a}_k$, which is applied to the current pose to obtain the desired next pose  $\hat{T}_{k+1}=T_k\hat{a}_k$. Because the evaluation replays recorded trajectories, an image is generally not available exactly at this pose. The acquisition is therefore queried for the recorded frame whose tracked pose is closest to the predicted one, and that frame becomes the next live observation; its recorded pose defines the state for the following iteration. Planning, pose update, and nearest-frame projection are repeated until a maximum of $N$ planning iterations is exceeded.

Using the closest recorded pose instead of the exact predicted pose is what makes retrospective evaluation possible in the first place, but it also introduces a degree of noise on the executed actions. If anything, this mirrors realistic deployment conditions, since a human operator likewise cannot execute a proposed action with perfect precision.

\subsubsection{Navigation planning}
We compare the proposed retrieval-based planner against representative strategies that differ in when, and how far ahead, the consequences of actions are assessed. UltraSeP-style guidance\cite{ultrasep} shapes its representations with predictive objectives during training, but at inference produces guidance through direct action prediction conditioned on previously executed actions and acquired images, without evaluating candidate actions online. Cardiac Copilot-style guidance\cite{cardiaccopilot} instead performs single-step inference-time planning, in which a learned forward model assesses the effect of a proposed action from the live image and refines the proposed actions, without look-ahead beyond a single step. The proposed method performs retrieval-based receding-horizon planning, with settings selected on the validation set: a horizon of $H=6$ and cost-function weights of: \(\lambda_{\mathrm{traj}} = 1\), \(\lambda_{\mathrm{mom}} = 0.3\), \(\lambda_{\mathrm{xstep}} = 0.2\).

We report two metrics. The navigation success rate is the fraction of episodes that reach the target view, reported per initialization type and per target view. The backtracking rate is the fraction of consecutive, nonzero action pairs whose cosine similarity is non-positive, reported separately for the translation and rotation components; zero (stay) actions are excluded from these pairs, and lower values indicate less backtracking.

\begin{figure}[t]
    \centering
    \includegraphics[width=0.99\columnwidth]{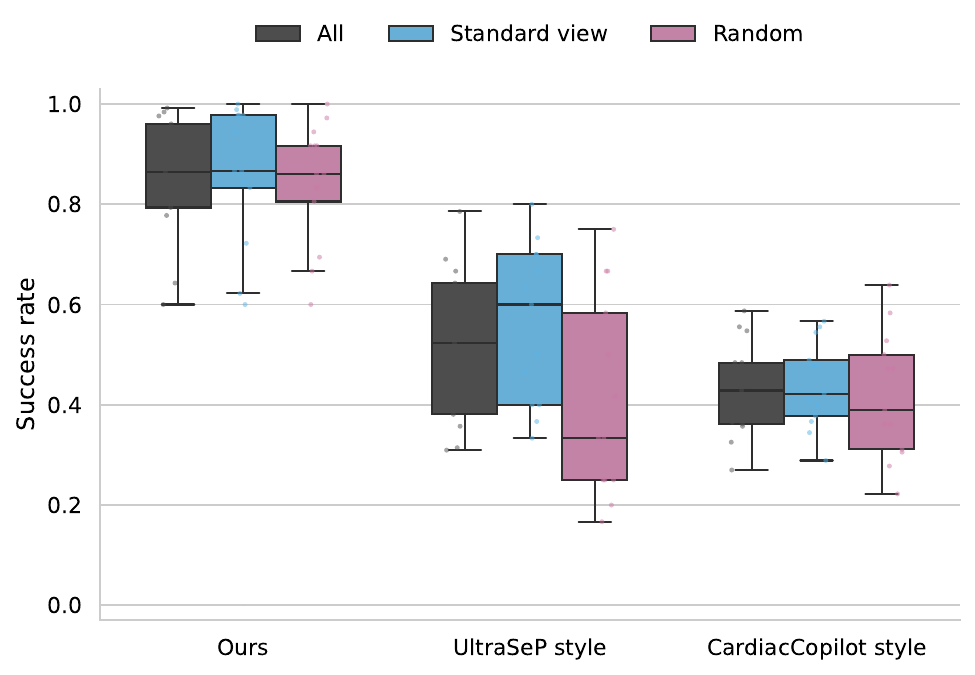}
    \caption{Success rates across test-set rollouts, separated by navigation initialization from a standard view and a random image. }
    \label{fig:successrate}
\end{figure}

\begin{figure}[t]
    \centering
    \includegraphics[width=0.99\columnwidth]{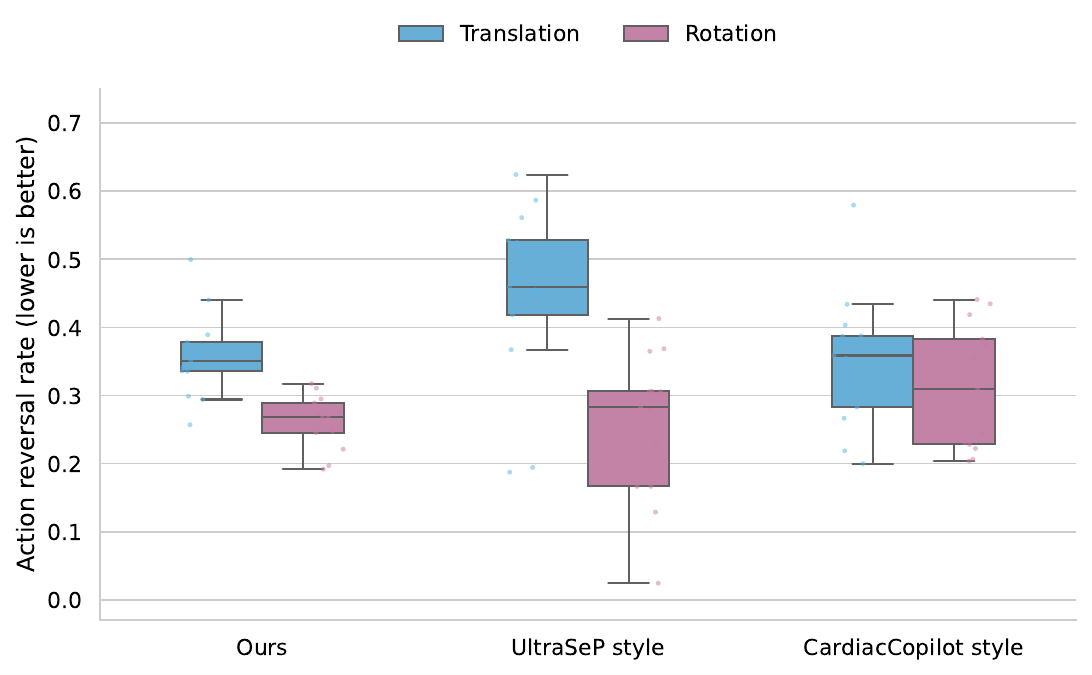}
    \caption{Fraction of consecutive nonzero global actions with non-positive cosine similarity, reported separately for translation and rotation, computed over successful episodes with more than two actions, up to the first arrival at the target view. Lower values indicate smoother trajectories with fewer direction reversals.}
    \label{fig:backtracking}
\end{figure}

\begin{figure}[t]
    \centering
    \includegraphics[width=0.99\columnwidth]{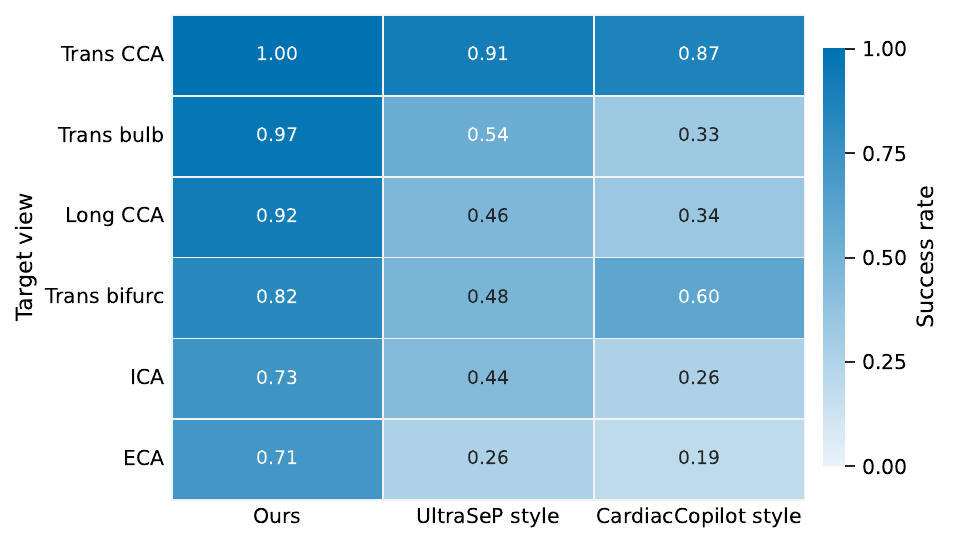}
    \caption{Success rate per target view, highlighting easier target views and more challenging ones, with our method having a higher success rate for all views.}
    \label{fig:view_performance}
\end{figure}

Fig.~\ref{fig:successrate} shows boxplots of the success rate achieved by the three methods. Across all test volunteers, target views, and initialization types, the proposed planner reaches the target view in 86\% of episodes, compared with 52\% for UltraSeP-style guidance and 43\% for Cardiac Copilot-style guidance. Random-frame initialization degrades success less for the proposed method than for the baselines, indicating lower sensitivity to the starting location. The per-view heatmap in Fig.~\ref{fig:view_performance} shows that this advantage is largest for the harder targets, the ECA and ICA, where the proposed method reaches the target view more often than the alternatives. Action smoothness of successful planning trajectories is reported in Fig.~\ref{fig:backtracking}: the three methods show similar rotation backtracking rates, while UltraSeP-style guidance has a higher translation backtracking rate than the other two methods.

\begin{figure}[t]
    \centering
    \includegraphics[width=0.99\columnwidth,trim={0.15cm 2.0cm 0.15cm 1.5cm}, clip]{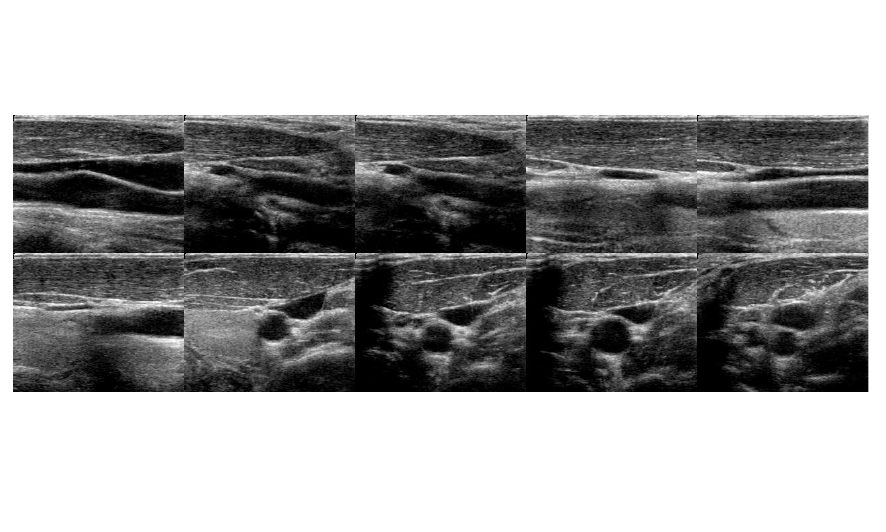}
    \caption{Example executed trajectory of offline planning, starting at the ICA navigating to the cross-sectional bifurcation.}
    \label{fig:navigation}
\end{figure}

Fig. \ref{fig:navigation} illustrates an example trajectory, navigating from the ICA to a transverse bifurcation using our proposed method.


\subsection{Ablation studies}
To quantify the contribution of the key components of the proposed navigation framework, we perform a series of ablation studies using the retrospective navigation benchmark on the validation set.

\paragraph{Planning horizon}
To evaluate the importance of multi-step planning, we vary the planning horizon from one to eight steps. The horizon-one setting corresponds to a greedy planner, while larger horizons enable progressively longer-term planning. As shown in Table~\ref{tab:depth_sweep}, moving from greedy to multi-step planning substantially improves the success rate, which then plateaus from \(H=5\) onwards. Although the greedy planner has the lowest rotational backtracking, it does so at a much lower success rate. Among the multi-step planners, backtracking is minimized around \(H=6\). We therefore select \(H=6\) as the optimal compromise between performance and computational cost, and use it in all other experiments.
\begin{table}[t]
\centering
\caption{Ablation: horizon sweep (validation set)}
\resizebox{0.99\linewidth}{!}{%
\begin{tabular}{lcccccccc}
\toprule
Horizon &
1 & 2 & 3 & 4 & 5 & 6 & 7 & 8 \\
\midrule
Success rate $\uparrow$ &
0.76 & 0.86 & 0.88 & 0.88 & 0.91 & 0.92 & 0.92 & \textbf{0.93} \\
Rot backtrack $\downarrow$ &
\textbf{0.22} & 0.29 & 0.28 & 0.27 & 0.28 & 0.26 & 0.26 & 0.27 \\
Trans backtrack $\downarrow$ &
0.37 & 0.41 & 0.38 & \textbf{0.35} & \textbf{0.35} & \textbf{0.35} & 0.37 & 0.37 \\
\bottomrule
\end{tabular}%
}
\label{tab:depth_sweep}
\end{table}

\paragraph{Cost function}
To evaluate the contribution of the individual trajectory scoring terms, we remove each component of the objective function while leaving the remaining terms unchanged. Table~\ref{tab:cost_ablation} shows that the terms play complementary roles: removing any term lowers the success rate, and removing \(J_{\mathrm{xstep}}\) additionally increases backtracking. The complete objective therefore provides the best balance between successful navigation and smooth trajectories.

\begin{table}[t]
\centering
\caption{Ablation cost function (validation set)}
\begin{tabular}{lcccc}
\toprule
Ablation &
Success rate $\uparrow$& Rot backtrack $\downarrow$& Trans backtrack $\downarrow$\\
\midrule
no \(J_{\mathrm{traj}}\)  & 0.89 & 0.24 & 0.31 \\
no \(J_{\mathrm{mom}}\)   & 0.84 & \textbf{0.22} & \textbf{0.29} \\
no \(J_{\mathrm{xstep}}\) & 0.90 & 0.34 & 0.39 \\
\midrule
Proposed                 & \textbf{0.92} & 0.26 & 0.35 \\
\bottomrule
\end{tabular}
\label{tab:cost_ablation}
\end{table}

\paragraph{Retrieval-induced transition model vs naive mimicking}
To evaluate the benefit of database-wide planning, we compare the proposed planner with a naive baseline that only follows temporal transitions within a single best-retrieved acquisition, using a fixed temporal stride. In contrast, the proposed planner repeatedly retrieves the most relevant states from the entire navigation database, allowing it to transition across acquisitions and volunteers whenever this leads to a more promising trajectory. Results in Table~\ref{tab:naive_retrieval} show that planning across the navigation database yields a higher success rate while simultaneously reducing backtracking, demonstrating the benefit of searching over multiple recorded trajectories rather than remaining within a single acquisition.

\begin{table}[t]
\centering
\caption{Naive database retrieval (validation set)}
\begin{tabular}{lccc}
\toprule
Stepsize &
Success rate $\uparrow$& Rot backtrack $\downarrow$& Trans backtrack $\downarrow$\\
\midrule
30  & 0.86 & 0.31 & 0.46 \\
80  & 0.82 & 0.34 & 0.44  \\
130 & 0.85 & 0.40 & 0.48  \\
\midrule
Proposed & \textbf{0.92} & \textbf{0.26} & \textbf{0.35} \\
\bottomrule
\end{tabular}
\label{tab:naive_retrieval}
\end{table}

\subsection{Navigation to previously acquired patient-specific view}
Unlike operator-guidance systems that navigate to predefined anatomical targets, our framework can target any view representable in latent space. We test this by returning to a patient-specific frame, replacing the target-view probability in the planner's cost with the cosine distance to that frame's latent; navigation is otherwise unchanged. 

A single goal frame is selected for each target view in each of the 13 held-out subjects, and the operator is given three attempts to reach it. Navigation starts within 500 frames of the target and at least 2 cm or 30° from it. Navigation runs for 20 planning steps and the performance is evaluated at the step with the smallest latent cosine distance to the target. This yields a success rate of 79\%. Among successful navigation attempts, the median euclidean translation error was 0.65 cm and the median geodesic rotation error was $4.5^\circ$. 

\subsection{Prospective feasibility study}
We performed a prospective feasibility study on three previously unseen healthy volunteers. The examinations were performed by the same operator who acquired the retrospective dataset. The navigation software was executed entirely on a standard CPU using a distilled version of the V-JEPA 2.1 model, with a planning horizon of three, a compromise between navigation performance and computational efficiency that enables real-time operation without GPU acceleration. The distilled model combines a MobileNetV3-Small image encoder with a small transformer that regresses the V-JEPA 2.1 patch tokens, reproducing the pooled navigation states at a cosine similarity of 0.97.

The operator followed on-screen guidance generated either by the proposed planner or by the Cardiac Copilot-inspired baseline. The active guidance method was randomized between navigation tasks, and the operator was blinded to the selected method. For each volunteer, both methods were evaluated twice, resulting in a total examination time of approximately 12 minutes per volunteer. A target view was considered reachable if it was successfully reached in at least one of the two attempts.

Each examination followed a fixed navigation protocol, sequentially guiding the operator through six standard views in the following order: short-axis CCA, short-axis carotid bulb, short-axis bifurcation, longitudinal CCA, longitudinal ICA, and longitudinal ECA. If the navigation method became disoriented, the operator restarted from the short-axis CCA view. A maximum of 30 seconds was allowed to reach each target view before proceeding to the next, and consecutive navigation tasks were separated by a 20-second pause.

To obtain ground-truth target images for later analysis, the dedicated view classifier (Section~\ref{sec:latent_representation}) continuously estimated the target-view probability during acquisition. Frames exceeding 0.85 were considered candidates, and the three highest-scoring candidates per target view were retained with five frames of context on either side for post-hoc selection of the most representative frame.

\begin{figure}[t]
    \centering
    \includegraphics[width=0.99\columnwidth,trim={0 1cm 0 0.9cm}, clip]{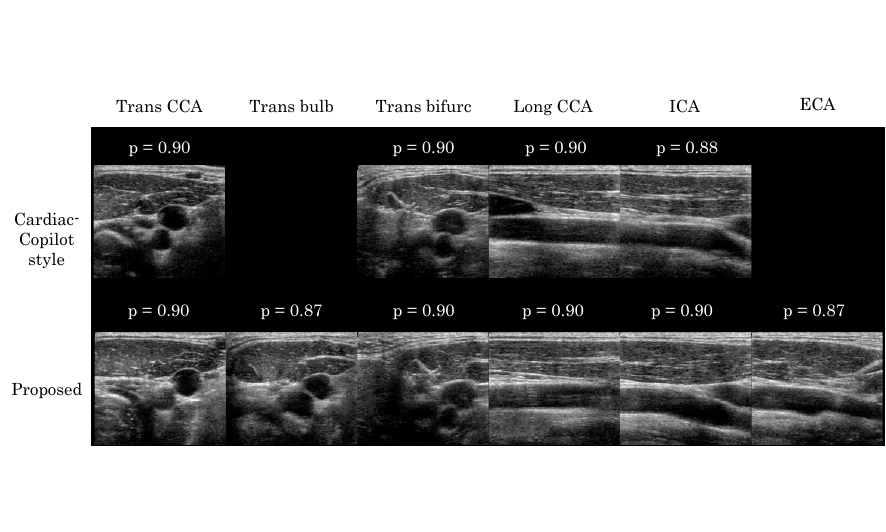}
    \caption{Acquisition results guided by both CardiacCopilot style method (top) and our proposed method (bottom). For views where no image is shown, the target was not reached. The view probabilities p indicate the view classification output for the corresponding target view.}
    \label{fig:lab_results}
\end{figure}

This resulted in a target-view reachability of 72\% for the Cardiac Copilot-inspired method and 83\% for the proposed method. An illustration of the target views is shown in Fig.~\ref{fig:lab_results}.

\section{Discussion}
In this work, we formulated ultrasound operator guidance as a receding-horizon planning problem in a learned world model and demonstrated that a non-parametric, retrieval-based transition model over previously recorded scans enables robust multi-step planning. On carotid ultrasound, the proposed approach consistently outperformed baselines in retrospective navigation experiments and remained effective in a prospective feasibility study on previously unseen volunteers, while operating in real time on a standard CPU. The framework also supports navigation to patient-specific target frames, highlighting its flexibility beyond fixed clinical protocols.

Several limitations should be acknowledged. First, the retrospective evaluation relies on recorded acquisitions, requiring each predicted probe pose to be projected onto the nearest recorded frame, introducing a coupling between the executed action and the available data. We note, however, that a comparable mismatch arises in real deployment, where the human operator cannot execute an action exactly. Second, future work should incorporate multi-rater annotation, enabling assessment of inter-rater agreement and view ambiguity. 

Third, data was acquired using a single ultrasound system, leaving robustness across scanners for future work. Likewise, the prospective proof-of-principle evaluation was limited to three healthy volunteers and a single operator. Future studies should include multiple operators, including novices. This poses an additional challenge: clear communication of guidance to the operator. While the planner predicts continuous six-degree-of-freedom probe motions, conveying these recommendations intuitively remains challenging, particularly when balancing the communication of large corrective motions together with smaller fine-adjustment cues.

Although this work focused on carotid ultrasound, the proposed planning framework is modality-agnostic and could be applied to other ultrasound guidance tasks. Interestingly, the learning-based baselines used for comparison were originally developed for cardiac ultrasound navigation, where anatomical variability and probe motions differ substantially from carotid imaging. Extending the proposed retrieval-based planning framework to cardiac ultrasound therefore represents a promising direction for future research.

\section*{Data Statement}
The data acquisition protocol was reviewed and approved by the ethical review board of Eindhoven University of Technology (Eindhoven, The Netherlands). All participants signed an informed consent form. 

\bibliographystyle{IEEEtran}
\bibliography{references}

\end{document}